\documentclass{article}
\usepackage{amsmath, amsfonts, sectsty, titling, algorithm, mathtools}
\usepackage{commath}
\usepackage[noend]{algpseudocode}
\usepackage{amssymb}
\usepackage{tikz}
\usetikzlibrary{bayesnet}
\usepackage{float}

\usepackage{caption}
\usepackage{subcaption}

\usepackage{etoolbox}
\usepackage{tikz}
\usetikzlibrary{tikzmark}
\usetikzlibrary{calc}

\newcommand{\N}{\mathcal{N}}
\newcommand{\E}{\mathbb{E}}
\newcommand\independent{\protect\mathpalette{\protect\independenT}{\perp}}
\def\independenT#1#2{\mathrel{\rlap{$#1#2$}\mkern2mu{#1#2}}}
\newcommand\numberthis{\addtocounter{equation}{1}\tag{\theequation}}
\DeclareMathOperator*{\argmax}{\arg\!\max}

\usepackage[final, nonatbib]{bdl_2018}

\usepackage[utf8]{inputenc} 
\usepackage[T1]{fontenc}    
\usepackage{hyperref}       
\usepackage{url}            
\usepackage{booktabs}       
\usepackage{amsfonts}       
\usepackage{nicefrac}       
\usepackage{microtype}      

\title{Deep Bayesian Nonparametric Factor Analysis}

\author{
  Arunesh Mittal  \\
  Columbia University\\
  \texttt{am4589@columbia.edu} \\ \And
  Paul Sajda \\
  Columbia University\\
  \texttt{psajda@columbia.edu} \\ \And
  John Paisley \\
  Columbia University\\
  \texttt{jpaisley@columbia.edu}
}

\begin{document}
\maketitle

\begin{abstract}
We propose a deep generative factor analysis model with beta process prior that can approximate complex non-factorial distributions over the latent codes. We outline a stochastic EM  algorithm for scalable inference in a specific instantiation of this model and present some preliminary results.

\end{abstract}

\section{Introduction}
Latent factor models provide a means to discover shared latent structure in large datasets by uncovering relationships between the observed data. These models make the assumption that the observed data is a mixture of latent factors. The data generating process for such models can be viewed as matrix factorization model, where the data matrix $X \in \mathbb{R}^{D\times N}$ is modelled as the matrix product $X \approx \Phi \xi$, where $\Phi \in \mathbb{R}^{D \times M}$ is the factor loading matrix and $\xi \in \mathbb{R}^{M \times N}$ is a matrix of latent variables. The columns $x_{1:N}$ of the matrix $X$ are the observed data, the columns $\Phi_{:,1:M}$ of $\Phi$ represent the $M$ factors, and the $M$ entries of each column $\xi_n$ of the latent matrix $\xi$ indicate the contribution of each of the factors that linearly combine to compose an observed data point $x_n$. That is, we allow each observation to possess combinations of up to $M$ latent features.  Then, given data the inference procedure entails learning the dictionary as well as the latent mixture components. For discrete $\xi$, computing the optimal values of $\xi_n$ requires a search over $2^M$ possible binary vectors and is computationally intractable even for modestly sized $M$, hence, we must resort to greedy search over this space. In addition, since $M$ is typically unknown, we would like to also infer the number of components in conjunction to the dictionary and components. 

Building upon previous work \cite{knowles2011nonparametric,paisley2009nonparametric,griffiths2011indian, zhou2012nonparametric, sertoglu2015scalable}, we propose a non-linear sparse coding factor analysis model based on Bayesian nonparametrics. The model employs non-linear ``multiplexer'' neural net that encodes latent binary vectors $z_n \in \{0,1\}^K$  to sparse latent variables $\xi_n \in \mathbb{R}_{+}^{M}$ . The network has the capacity to non-linearly explore the large parameter space over the latent encodings $\xi_n$. In addition, given a factorial distribution over $z_n$, the network can learn the correlation structure and approximate non-factorial distributions over the latent codes $\xi_n$ at the deepest layer. This allows the model to generate better samples than traditional linear factor models with factorial prior over the latents. By defining a sparse beta-Bernoulli process prior on the $z_n$, the model learns the optimal size $K$. Despite the non-linearity of the model, the parameters of the model are still interpretable as the interaction between the factor loading matrix $\Phi$ and the outputs of the multiplexer network $\xi_n$ is a linear operation. We propose a stochastic MAP-EM algorithm with a ``selective'' M step for efficient scalable inference in this model. \cite{sertoglu2015scalable}.

\section{Generative Model} \label{generative_model}
We use the finite limit approximation to the beta process \cite{griffiths2011indian, paisley2009nonparametric}. For $\pi_k \sim \mathrm{Beta}\left(\alpha \gamma / K, \alpha \left( 1- \gamma\ K \right)\right)$ and
$\Phi_k \sim g(\cdot)$, the random measure $H_K = \sum_{k=1}^{K} \pi_k \delta_{\Phi_k}$, $\lim_{K\to\infty} H_K$ converges in distribution to $H \sim BP(\alpha, \gamma)$. In addition, for $z_{nk} \sim \mathrm{Bern}(\pi_k)$, where $\pi_k \sim \mathrm{Beta}\left(\alpha \gamma / K, \alpha \left( 1- \gamma\ K \right)\right)$, $k= 1,\ldots,K$, the random measure $G_{n}^{K} = \sum_{k=1}^{K} z_{nk} \delta_{\Phi_k}$, $\lim_{K\to\infty} G_{n}^{K}$ converges to a Bernoulli process \cite{paisley2016constructive}.

We model the generative process for the data as follows: Given data $x = \{x_n\}_{n=1}^{N}$, the corresponding latent factors $z_n \in \{z_n\}_{n=1}^{N}$ are drawn from a Bernoulli process (BeP) parameterized by a beta process (BP), where, the Bernoulli process prior over each of the $k$ factors $z_{nk} \in z_n$, is parameterized by $\pi_k$ drawn from a beta process. Each latent variable $\xi_n$ is drawn from the distribution $h(\cdot)$ parameterized by the latent factor $z_n$ via a neural network $\varphi(\cdot)$ with parameters $\theta_{\varphi}$, where, the $L$ layered neural network $\varphi(\cdot)$, maps the binary latent code $z_n \in \{0,1\}^K$ to $\mathcal{S} \subset \mathbb{R}^M$ via a neural net: $\varphi(z_n) = \sigma_{L}W_L\sigma(W_{L-1} \sigma_{l-1}( \ldots  W_{l} \sigma_l( \ldots \sigma_0(W_0 z_n))))$. The factor loading matrix $\Phi \in \{\mathcal{M} : \mathcal{M} \subset \mathbb{R}^{D \times M}\}$ is drawn from the distribution $g(\cdot)$ with parameters $\theta_{\Phi}$.  The scaling factor $\lambda_n$ for each data point $x_n$ is drawn from a Gaussian distribution. Finally, each data point $x_n$ is drawn from a isotropic Gaussian distribution where the mean is parameterized by the matrix-vector product of factor loading matrix $\Phi$ and the encoded latent factor $\xi_n$, scaled by $\lambda_n$. 

\noindent
 \begin{alignat*}{4}
    &\pi_k     &&\sim  \mathrm{Beta}\left(\alpha (\gamma / K), \alpha \left( 1- \gamma\ K \right)\right) \\
    &z_{nk} &&\sim \mathrm{Bern}(\pi_k)\\  
    &\xi_n &&\sim h(\varphi(z_n; \theta_{\xi})) \\
    &\Phi &&\sim g(\Phi \ ; \theta_{\Phi}) \\
    &\lambda_{n} &&\sim \N(0, c)  \\
    &x_n       &&\sim \N(\lambda_n \cdot \Phi \ \xi, \sigma^{2} I) 
\end{alignat*}

In the subsequent discussion, we choose specific distributional forms for $h(\cdot)$ and $g(\cdot)$ to illustrate an instance of the general generative model outlined above. We illustrate how the generative model with  sparsity inducing beta prior can be applied to non-parametric dictionary learning. Specifically, for dictionary learning, we choose $h(\cdot)$ to be a $M$-dimensional Dirichlet distribution $\mathrm{Dir}(\xi_n \vert \varphi(z_{n1}), \ldots, \varphi(z_{nK}) )$ and choose  $g(\cdot)$ to be the Dirac measure $\delta_{\theta_{\Phi}}(\Phi)$, where $\theta_{\Phi} = D \in \mathbb{R}^{D \times M}$. Each data point $x_n$ is then drawn from $\N(\lambda_n \cdot f_{\theta}(z_n), \sigma^{2} I)$, where, $f_{\theta}(z_n) = \Phi \ \mathbb{E}[\xi_n]$ and $\theta = \{\theta_{\xi}, \theta_{\Phi}\}$. For our preliminary experiments we chose a dense neural network that maps $z_n$ to $\mathbb{E}[\xi_n]$, where the last layer was a softmax layer that output values over the $\Delta_{M-1}$ simplex. More generally, one could use $\varphi(\cdot)$ to parameterize the natural parameters of a exponential family distribution such as a gamma or Poisson distribution \cite{ranganath2015deep}.

\section{MAP-EM Inference}
We propose a MAP-EM algorithm to perform inference in this model. We compute point estimates for $z$ and $\theta$ and posterior distributions over $\pi$ and $\lambda$. Since, $\pi  \independent \lambda \mid z$, the conditional posterior distribution factorizes as:  $p(\pi, \lambda \mid x, z, \theta) = p(\pi \mid z)p(\lambda \mid x, \theta)$. Hence, exploiting conjugacy in the model, we can analytically compute posterior distributions $q(\pi)$ and $q(\lambda)$. We compute point estimates for $z$ and $\theta$.

\subsection{Stochastic E-Step}
By conjugacy of the beta to the Bernoulli process we can compute $q(\pi) \triangleq  p(\pi \mid z)$ analytically. In addition, to make inference scalable, we employ stochastic inference for $\pi_k$ and use natural gradient of the posterior parameters $(a_k , b_k)$ using random batch of data $S \subset \{x_n\}_{n=1}^{N}$to update the posterior parameters , where $\eta$ is the stochastic gradient step \cite{hoffman2013stochastic}: 
\begin{align*}
        q(\pi) = \textstyle\prod\nolimits_{k} \mathrm{Beta}(\pi_k \mid a_k, b_k); \quad &a_k^{\prime} = \alpha \textstyle\frac{\gamma}{K} + \frac{N}{|S|}\sum\nolimits_{n \in S} z_{nk} \ ;   b_k^{\prime} = \alpha \left(1-\frac{\gamma}{K}\right) + \frac{N}{|S|} \sum_{n \in S} 
                       \left(1-z_{nk}\right) \\
                                       &a_k \gets (1-\eta) a_k + \eta \ a_k^{\prime}  \ \ \quad
                            ;b_k \gets (1-\eta) b_k + \eta \ b_k^{\prime} \numberthis{} \label{eq:q_pi_ab}
\end{align*}
In addition, since $q(\lambda) \triangleq p(\lambda \mid x, \theta, z)$ factorizes as $\prod_n q(\lambda_n) = \prod_n p(\lambda_n \mid x_n, \theta, z_n)$, and the posterior distribution over $\lambda_n$ is also a Gaussian, we can analytically compute the posterior $q(\lambda_n)$:

\noindent\begin{minipage}{.4\linewidth}
    \begin{align*}
        q(\lambda_n) 
        &= \N(\lambda_n \mid \mu_{\lambda_n \mid x_n,  z_n, \theta}, \sigma^{2}_{\lambda_n \mid x_n, z_n, \theta})
    \end{align*}
    
\end{minipage}
\begin{minipage}{.6\linewidth}
    \begin{align*}
        \sigma^{2}_{\lambda_n \mid x_n, z_n, \theta} &= \left( c^{-1} + f_\theta(z_n)^{\top} f_\theta(z_n) / \sigma^2 \right)^{-1} \numberthis{} \label{eq:q_lambda_mu} \\
        \mu_{\lambda_n \mid x_n, z_n, \theta}        &= (\sigma^{2}_{\lambda_n \mid x_n, z_n,\theta})
                                                     (f_\theta(z_n)^{\top} x_n)/ \sigma^2 \numberthis{} \label{eq:q_lambda_sigmasq}
    \end{align*}
\end{minipage}

\renewcommand{\thealgorithm}{}
\newcommand*{\Break}{\textbf{break}}
\algnewcommand{\And}{\textbf{and}}
\algnewcommand{\algorithmicgoto}{\textbf{go to}}%
\algnewcommand{\Goto}[1]{\algorithmicgoto~\ref{#1}}%

\errorcontextlines\maxdimen

\newcommand{\ALGtikzmarkcolor}{black}
\newcommand{\ALGtikzmarkextraindent}{4pt}
\newcommand{\ALGtikzmarkverticaloffsetstart}{-.5ex}
\newcommand{\ALGtikzmarkverticaloffsetend}{-.5ex}
\makeatletter
\newcounter{ALG@tikzmark@tempcnta}

\newcommand\ALG@tikzmark@start{%
    \global\let\ALG@tikzmark@last\ALG@tikzmark@starttext%
    \expandafter\edef\csname ALG@tikzmark@\theALG@nested\endcsname{\theALG@tikzmark@tempcnta}%
    \tikzmark{ALG@tikzmark@start@\csname ALG@tikzmark@\theALG@nested\endcsname}%
    \addtocounter{ALG@tikzmark@tempcnta}{1}%
}

\def\ALG@tikzmark@starttext{start}
\newcommand\ALG@tikzmark@end{%
    \ifx\ALG@tikzmark@last\ALG@tikzmark@starttext
    \else
        \tikzmark{ALG@tikzmark@end@\csname ALG@tikzmark@\theALG@nested\endcsname}%
        \tikz[overlay,remember picture] \draw[\ALGtikzmarkcolor] let \p{S}=($(pic cs:ALG@tikzmark@start@\csname ALG@tikzmark@\theALG@nested\endcsname)+(\ALGtikzmarkextraindent,\ALGtikzmarkverticaloffsetstart)$), \p{E}=($(pic cs:ALG@tikzmark@end@\csname ALG@tikzmark@\theALG@nested\endcsname)+(\ALGtikzmarkextraindent,\ALGtikzmarkverticaloffsetend)$) in (\x{S},\y{S})--(\x{S},\y{E});%
    \fi
    \gdef\ALG@tikzmark@last{end}%
}

\apptocmd{\ALG@beginblock}{\ALG@tikzmark@start}{}{\errmessage{failed to patch}}
\pretocmd{\ALG@endblock}{\ALG@tikzmark@end}{}{\errmessage{failed to patch}}
\makeatother

\makeatletter
\renewcommand{\ALG@name}{Algorithm 1:}
\makeatother

\makeatletter
\expandafter\patchcmd\csname\string\algorithmic\endcsname{\itemsep\z@}{\itemsep=0ex}{}{} 
\makeatother

\makeatletter
\renewcommand{\ALG@beginalgorithmic}{\footnotesize}
\makeatother

\begin{algorithm}
\caption{Stochastic MAP-EM for Sparse Coding}
\begin{algorithmic}[t!]
    
    \Function{sparse\_code}{$x \in \mathbb{R}^{N \times D}, z \in \{0,1\}^{N \times K}$} 
    \label{alg:sparse_code}    
    \State \textbf{Initialize} $\ \rho, \ a_k, \ b_k, \ \kappa \in (0.5, 1], \ \tau_0 \geq 0$ 
        
    \While{\textbf{not} \text{converged}}
        \State $S_t \subset \{1, \ldots, N\}$   \Comment{Take mini-batch}
        \State $\eta \gets (\tau_0 + t)^{-\kappa} $ \Comment{Learning rate schedule}
        \For{$n \in S_t$}    \Comment{Update $q(\lambda)$ using Eq. \eqref{eq:q_lambda_mu} and \eqref{eq:q_lambda_sigmasq}}
            \State $q(\lambda_n) := \N(\mu_{\lambda_n \mid x_n, z_n, \theta}, \sigma_{\lambda_n \mid x_n, z_n, \theta})$
        \EndFor
        
        \For{$n \in S_t$}   \Comment{Update $q(\pi)$ using Eq. \eqref{eq:q_pi_ab}}
            \For{$k \in \{1 \ldots K\}$}            
                \State $q(\pi_k)  := \mathrm{Beta}(\pi_k \mid a_k, b_k) $
            \EndFor
        \EndFor
        
        \State \textbf{Initialize:} $\forall \, (n \in S_t) \ \Omega_n = \emptyset$ 
        \For{$n \in S_t$}  \label{marker}  \Comment{Update z to maximize $\mathcal{L}_{\pi}(z,\theta)$ using Eq. \eqref{eq:p_x_sc} and \eqref{eq:p_z}} 
            \State \textbf{Initialize:} $\forall \, k  \ z_{nk} = 0,  \ \zeta^- = 0$
            \While{$\| \Omega_n \|_{0} < L$} 
                \State $j^* \gets \argmax_{j \setminus \Omega_n} \ \ln p(x_n \mid z_{\Omega_n} = 1, z_{nj} = 1) + 
                        \E_{q(\pi)}\left[\ln p(z_{\Omega_n} = 1, z_{nj} = 1 \mid \pi_{k})\right]$ 
                \State $\zeta^+ = \ln p(x_n \mid z_{\Omega_n} = 1, z_{nj*} = 1) + 
                     \E_{q(\pi)}\left[\ln p(z_{\Omega_n} = 1, z_{nj^*} = 1 \mid \pi_{k})\right]$
                \If{$\zeta^{+} > \zeta^{-}$} 
                    \State $\Omega_n  \gets \Omega_n \cup \{ j \}$
                    \State $\zeta^{-} \gets \zeta^{+}$
                \Else
                    \State \Break
                \EndIf
            \EndWhile
        \EndFor
        \State $\theta \gets \textsc{adam}(\E_{q(\lambda)}[\ln p(x \mid \theta, z, \lambda)], \ \text{stepsize} = \rho)$ \Comment{Update $\theta$ to max. $\mathcal{L}_{\pi, \lambda}(z, \theta)$ using Eq. \eqref{eq:p_x_nn}}
    \EndWhile
    \EndFunction
\end{algorithmic}
\end{algorithm}

\subsection{M-Step}
Given the joint density, we can compute the MAP objective $\mathcal{L}_{\lambda, \pi}(z,\theta)$, alternatively, we marginalize out $\lambda$ from $p(x,\lambda \mid \theta, z)$ to compute the objective $\mathcal{L}_{\pi}(z,\theta)$: 
\begin{align*}
    \mathcal{L}_{\lambda, \pi}(z,\theta) &= \E_{q(\lambda)} \left[ \ln p(x \mid \theta, z, \lambda) \right] + \E_{q(\pi)}\left[\ln 
                            p(z \mid \pi) \right] \\
    \mathcal{L}_{\pi}(z,\theta) &=  \ln p(x \mid \theta, z) + \E_{q(\pi)}\left[\ln 
                            p(z \mid \pi) \right] 
\end{align*}
The marginal over $x_n$ and the conditional expectations can be computed analytically:
\begin{align*}
&\ln p(x_n \mid \theta, z_n) = -.5  \big[ \ln \left(
                                1 +  c \ \sigma^{-2} f_{\theta}(z_n)^{\top} f_{\theta}(z_n) 
                                \right) + \ln  \left| \sigma^2 I  \right|  + \\ 
                                &\hspace{10em} x_n^{\top} \left( \sigma^{-2}I - \sigma^{-2} f_\theta(z_n)f_\theta(z_n)^{\top} / c^{-1}\sigma^2 + f_\theta(z_n)^{\top}f_\theta(z_n)\right) x_n \big]\numberthis{} \label{eq:p_x_sc} \\
&\E_{q(\lambda)} \left[ \ln p(x \mid \theta, z, \lambda) \right] =  \| x_n -  \mu_{\lambda_n \mid x_n, z_n, \theta} \cdot f_\theta(z_n) \|^2 
                                                             + \sigma^{2}_{\lambda_n \mid x_n, z_n, \theta} \cdot f_\theta(z_n)^\top f_\theta(z_n) \numberthis{} \label{eq:p_x_nn}\\
&\E_{q(\pi)}\left[\ln p(z_n \mid \pi) \right] = \textstyle\sum\nolimits_{k} z_{nk} [\psi(a_k - \psi(a_k + b_k)] + (1-z_{nk}) [\psi(b_k) -                                            \psi(a_k + b_k)] \numberthis{} \label{eq:p_z}
\end{align*}
where $\psi(\cdot)$ is the digamma function.

To optimize $\mathcal{L}_{\lambda, \pi}(z,\theta)$, we employ a greedy algorithm, which is similar to the matching pursuit used by K-SVD \cite{aharon2006k}. We use $z_{\Omega_n}$ to denote a $k$-vector, corresponding to the latent vector for the $n$th data point, where,  $\forall j \in \Omega_n, z_{nj} = 1$ and $\forall j \not\in \Omega_n,  z_{nj} = 0$. To compute the sparse code given a data point $x_n$, we start with an empty active set $\Omega_n$,  then $\forall j \in \{1,\ldots, K\}$, we individually set each $z_{nj} = 1$ to find $j^* \in \{1,\ldots,K\} \setminus \Omega_n$ that maximizes $\mathcal{L}_{\lambda, \pi}(z_{\Omega_n \cup \{j^*\}},\theta)$. We compute the scores $\zeta^{+} \triangleq \mathcal{L}_{\lambda, \pi}(z_{\Omega_n \cup \{j^*\}},\theta)$ and
$\zeta^{-} \triangleq \mathcal{L}_{\lambda, \pi}(z_{\Omega_n},\theta)$. We add  $j^*$ to $\Omega_n$ only if $\zeta^{+} > \zeta^{-}$, this step is necessary because unlike matching pursuit, the neural net $\varphi(\cdot)$ introduces a non-linearity from $z_{\Omega_n}$ to $\xi_n$, hence, adding  $j^*$ to $\Omega_n$ can decrease  $\mathcal{L}_{\lambda, \pi}(z_{\Omega_n},\theta)$. For each $x_n$, we repeat the preceding greedy steps to sequentially add factors to $\Omega_n$ till $\mathcal{L}_{\lambda, \pi}(z_{\Omega_n},\theta)$ ceases to monotonically increase. 

For optimization of  $\mathcal{L}_{\lambda, \pi}(z,\theta)$ w.r.t. $z$, the scoring procedure to add factors to $\Omega_n$ is similar to the correlation score used by matching pursuit step used K-SVD \cite{aharon2006k}. The expected log prior on $z$ imposes an approximate beta process penalty. Low probability factors learned through $q(\pi)$ will lead negative scores $\zeta^{+}$, and hence eliminate latent factors to encourage sparsity of $z_n$. During optimization once  $q(\pi_k)$ falls below a certain threshold we no longer need to consider the $z_{nk}$ when optimizing $\mathcal{L}_{\lambda, \pi}(z,\theta)$. This allows for speed up of the sparse coding routine over iterations.

We can maximize $\mathcal{L}_{\pi}(z,\theta)$ w.r.t. $\theta$, which includes both the neural net parameters $\theta_{\xi}$ and the dictionary $\theta_{\Phi}$ by stochastic optimization using ADAM \cite{kingma2014adam}. First order gradient methods with moment estimates such as ADAM, can implicitly take into account the rate of change of natural parameters $(a_k , b_k)$ for $q(\pi)$ when optimizing the neural net parameters $\theta_{\xi}$ and the dictionary $\theta_{\Phi}$. The full sparse coding algorithm is outlined in Alg. (\ref{alg:sparse_code}).

\section{Preliminary Results}
\begin{figure}
\begin{subfigure}[t]{.5\textwidth}\centering
  \includegraphics[width=1\linewidth]{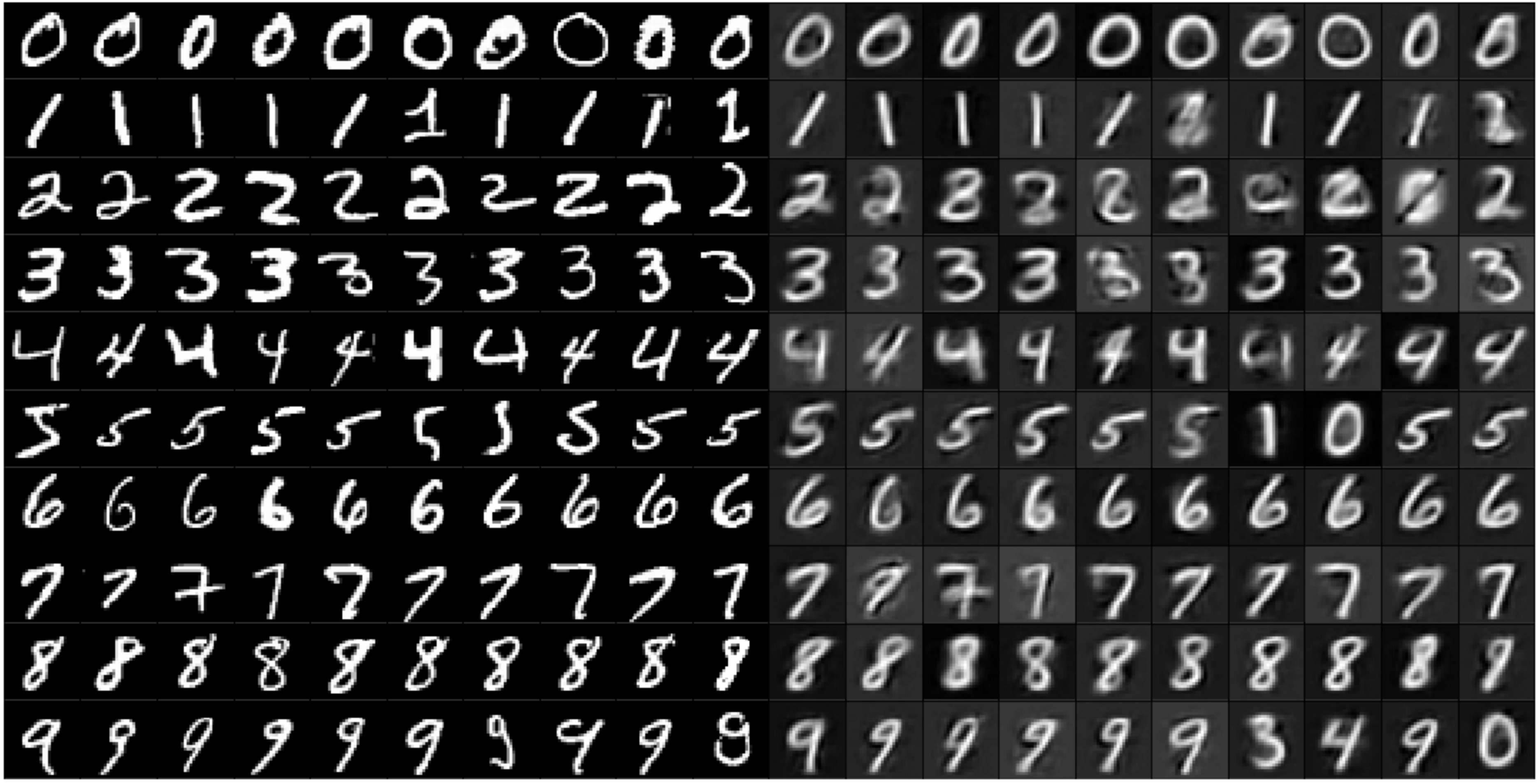}
  \caption{Example reconstructions of MNIST digits from inferred sparse codes. Left 10 columns true data, right 10 columns approximations.}
  \label{fig:truth_v_approx_mnist}
\end{subfigure}%
\begin{subfigure}[t]{.5\textwidth}\centering
  \includegraphics[width=.5\linewidth]{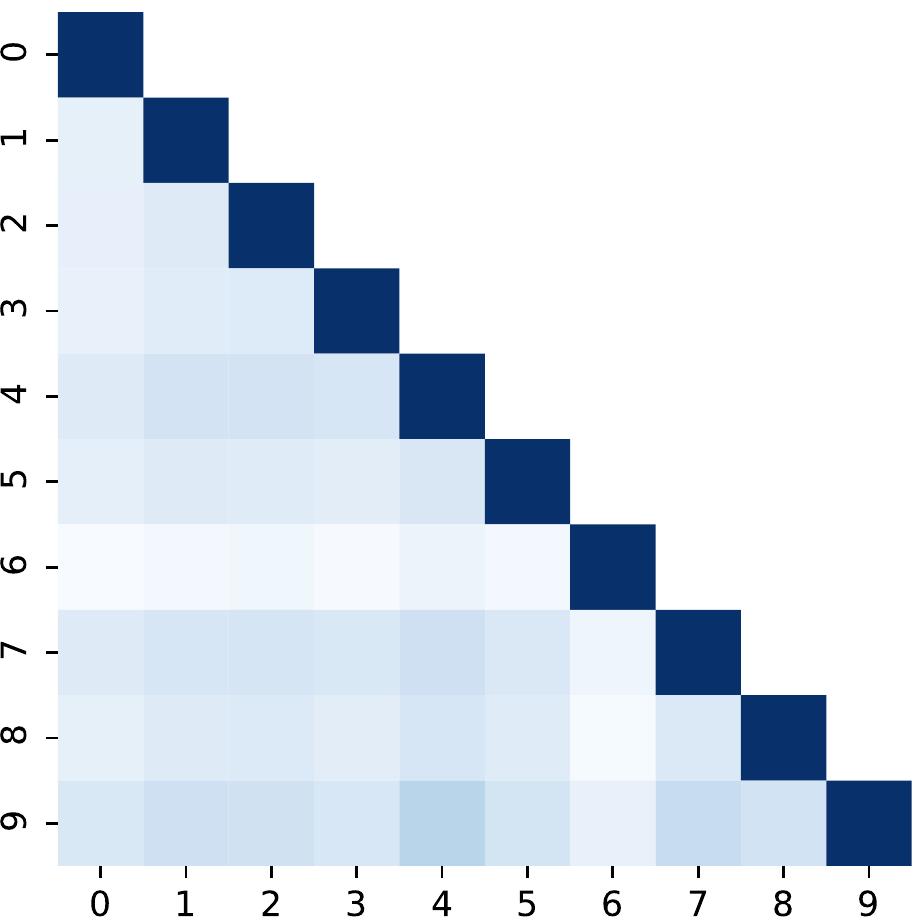}
  \caption{Expected factor  sharing between pairs of digits.}
  \label{fig:factor_sharing}
\end{subfigure}
\caption{}
\end{figure}

\begin{figure}[h]
    \centering
    \includegraphics[width=1\textwidth]{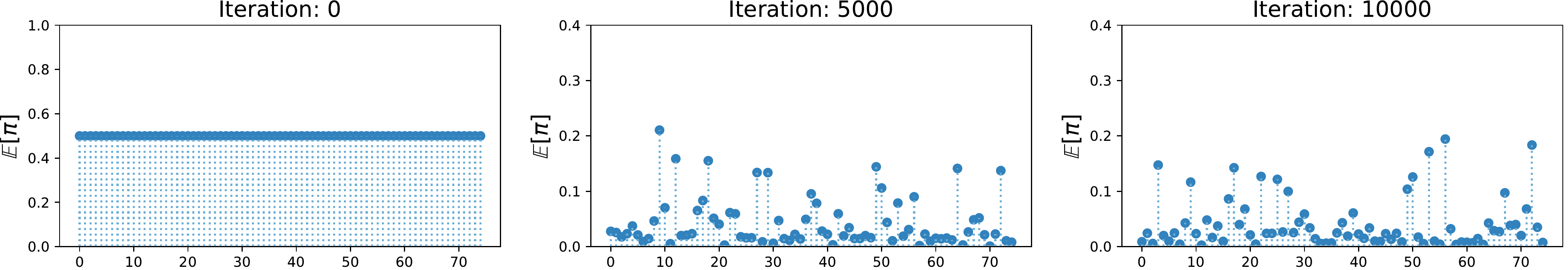}
    \caption{Inferred $\E[\pi]$ indicating increasing sparsity over iterations.}
    \label{fig:E_pi_over_iterations}
\end{figure}

We trained a $3$ layered neural network $\varphi(\cdot)$ with $100$ hidden units  with a softmax output layer. We chose a factor loading matrix $D$ of size $78 \times 256$ and set the number of factors $K = 75$. For our prior parameters we set $\alpha =1$,$\gamma = 1$, $\sigma=10$ and $c=1e15$. We set a constant learning rate for the neural network to $0.001$ and the learning rate schedule parameters $\tau = 100$ and $\kappa = 0.6$ for $q(\pi)$ . We trained our model on the MNIST data with batch size of $200$ for $10,000$ iterations. The deep nonparametric dictionary learning model was able to reconstruct digits well from the inferred sparse codes $z_n$ Fig. (\ref{fig:truth_v_approx_mnist}). Over the course of training, the beta-process sparsity prior encouraged only a small subset of the $K$ factors to be ultimately used Fig. (\ref{fig:E_pi_over_iterations}) while optimizing the factor loading matrix as well as the neural net parameters.  In addition, the model learned shared factors across the digits. We show the factor sharing across the digits by calculating the expected number of factors shared between all pairs of two digits (normalized by the largest value)  Fig. (\ref{fig:factor_sharing}) .

\begin{figure}[h]
    \centering
    \includegraphics[width=1\textwidth]{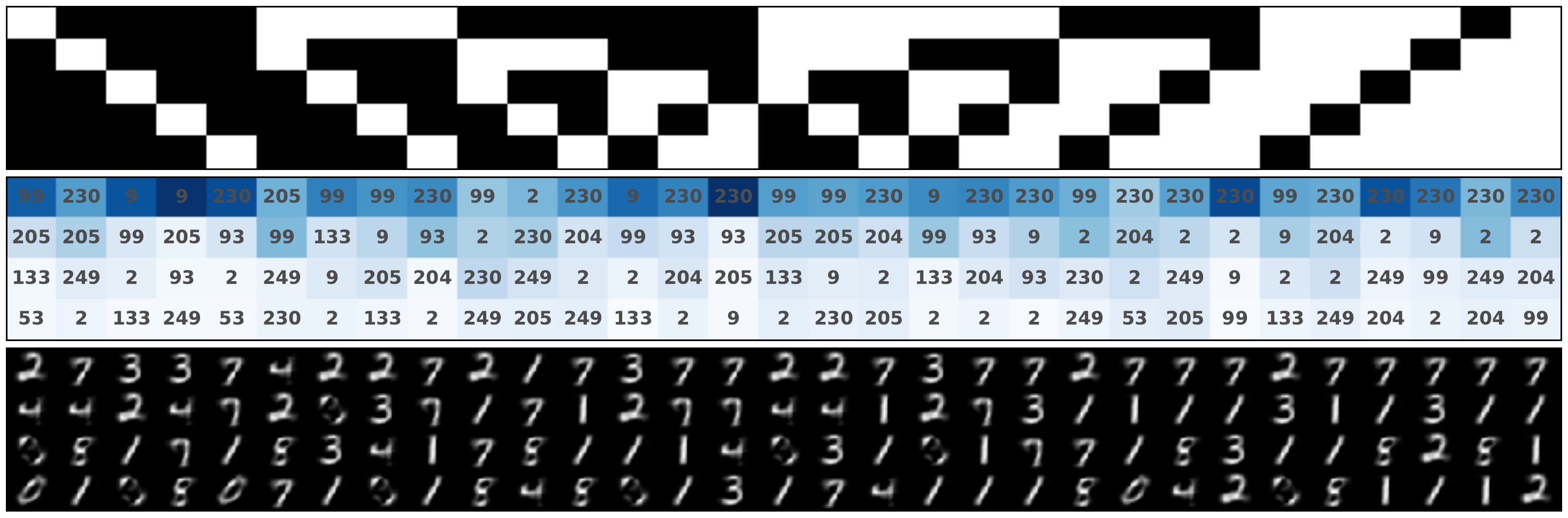}
    \caption{Combinatorial outputs for top five input bits with highest $\E[\pi]$. (Top) Each column  corresponds to input sequence $z_n$ and each row entry   indicates whether the top $k$th element of sequence was selected or not (white = 1/black = 0). (Middle) The corresponding output $\varphi(z_n; \theta_{\xi})$, where each column corresponds to the top four output elements with highest activation probability for $z_n$th input. The numbers corresponds to the output element index, and the color indicates the probability of a particular output element being turned on given an input sequence $z_n$. (Bottom) Inferred factor loadings corresponding to the top four output elements with highest activation probability. }
    \label{fig:i_o_bits}
\end{figure}  
 
The non-linear factor analysis model is a more expressive model compared to a linear factor analysis model due to the fact that the neural net $\varphi(z_n; \theta_{\xi})$ can can index the factor loading matrix $\Phi$ in a non-linear fashion. That is, indices selected by a factor sequence $z_n$ such that $z_{nj}=1$ can be unselected by a sequence $z_m$ with $z_{mj}=1, z_{ml}=1$. To illustrate this we trained the same model as above with the same training parameters, however, constrained the factor loading matrix entries to be non-negative. This model can be viewed as non-linear non-negative matrix factorization. We then illustrate the non-linearity of factor selection in Fig. (\ref{fig:i_o_bits}), where the difference between factor sequences between column one and six is just on additional factor, however, the top five factor loadings selected by the network, given the two sequences, are entirely different. 

\section{Conclusion and Future Work}
Our non-linear factor analysis model can approximate complex non-factorial distributions over the latent codes. Our MAP-EM algorithm, allows for the exploration of the large combinatorial space over the latent encodings. In addition, the beta-process sparsity prior encourages only a small subset of the $K$ factors to be utilized. In implementations of our our algorithm, one could choose to start ignoring indices of factor sequences whose expected selection falls below certain threshold. This allows our inference procedure to speed up over time as we have to check a smaller number of indices during the M step. 

Our specific algorithm for deep sparse coding leverages conjugacy in the model to marginalize out the scaling factor $\lambda$ during the M step. This makes the algorithm resilient to scaling of the data since we account for scaling of each data point $x_n$ by inferring the scaling factor $\lambda_n$. As discussed, deep sparse coding for dictionary learning is a specific instantiation of the generative model outlined in section \ref{generative_model}. More generally, for appropriate choice of priors $g(\cdot)$ on $\Phi$, and $h(\cdot)$ of $\xi$, the generative model encompasses a broad class of models such as PCA, sparse coding, sparse PCA and sparse matrix factorization/non-negative sparse coding. In future work, we plan to extend the above inference algorithm to this broader class of models. 


\begin{thebibliography}{10}

\bibitem{aharon2006k}
M.~Aharon, M.~Elad, A.~Bruckstein, et~al.
\newblock K-svd: An algorithm for designing overcomplete dictionaries for
  sparse representation.
\newblock {\em IEEE Transactions on signal processing}, 54(11):4311, 2006.

\bibitem{griffiths2011indian}
T.~L. Griffiths and Z.~Ghahramani.
\newblock The indian buffet process: An introduction and review.
\newblock {\em Journal of Machine Learning Research}, 12(Apr):1185--1224, 2011.

\bibitem{hoffman2013stochastic}
M.~D. Hoffman, D.~M. Blei, C.~Wang, and J.~Paisley.
\newblock Stochastic variational inference.
\newblock {\em The Journal of Machine Learning Research}, 14(1):1303--1347,
  2013.

\bibitem{kingma2014adam}
D.~P. Kingma and J.~Ba.
\newblock Adam: A method for stochastic optimization.
\newblock {\em arXiv preprint arXiv:1412.6980}, 2014.

\bibitem{knowles2011nonparametric}
D.~Knowles and Z.~Ghahramani.
\newblock Nonparametric bayesian sparse factor models with application to gene
  expression modeling.
\newblock {\em The Annals of Applied Statistics}, pages 1534--1552, 2011.

\bibitem{paisley2009nonparametric}
J.~Paisley and L.~Carin.
\newblock Nonparametric factor analysis with beta process priors.
\newblock In {\em Proceedings of the 26th Annual International Conference on
  Machine Learning}, pages 777--784. ACM, 2009.

\bibitem{paisley2016constructive}
J.~Paisley and M.~I. Jordan.
\newblock A constructive definition of the beta process.
\newblock {\em arXiv preprint arXiv:1604.00685}, 2016.

\bibitem{ranganath2015deep}
R.~Ranganath, L.~Tang, L.~Charlin, and D.~Blei.
\newblock Deep exponential families.
\newblock In {\em Artificial Intelligence and Statistics}, pages 762--771,
  2015.

\bibitem{sertoglu2015scalable}
S.~Sertoglu and J.~Paisley.
\newblock Scalable bayesian nonparametric dictionary learning.
\newblock In {\em EUSIPCO}, pages 2771--2775, 2015.

\bibitem{zhou2012nonparametric}
M.~Zhou, H.~Chen, J.~Paisley, L.~Ren, L.~Li, Z.~Xing, D.~Dunson, G.~Sapiro, and
  L.~Carin.
\newblock Nonparametric bayesian dictionary learning for analysis of noisy and
  incomplete images.
\newblock {\em IEEE Transactions on Image Processing}, 21(1):130--144, 2012.

\end{thebibliography}

\end{document}